\documentclass[runningheads]{llncs}

\usepackage{iciap}

\usepackage{iciapabbrv}

\usepackage{graphicx}
\usepackage{booktabs}

\usepackage[accsupp]{axessibility}  

\usepackage{hyperref}

\usepackage{orcidlink}

\usepackage{cite}
\usepackage{amsmath,amssymb,amsfonts}
\usepackage{algorithmic}
\usepackage{graphicx}
\usepackage{textcomp}
\usepackage{xcolor}
\def\BibTeX{{\rm B\kern-.05em{\sc i\kern-.025em b}\kern-.08em
    T\kern-.1667em\lower.7ex\hbox{E}\kern-.125emX}}

\usepackage{mathtools}
\DeclarePairedDelimiterX{\infdivx}[2]{(}{)}{%
  #1\;\delimsize\|\;#2%
}
\newcommand{\infdiv}{D_{KL}\infdivx}

\DeclareUnicodeCharacter{2212}{-}

\begin{document}

\title{VidCLearn: A Continual Learning Approach for Text-to-Video Generation} 


\author{Luca Zanchetta\inst{1}\orcidlink{0000-1111-2222-3333} \and
Lorenzo Papa\inst{2}\orcidlink{1111-2222-3333-4444} \and
Luca Maiano\inst{1}\orcidlink{2222--3333-4444-5555} \and 
Irene Amerini \inst{1}\orcidlink{2222--3333-4444-5555}
}

\authorrunning{L.~Zanchetta et al.}

\institute{Sapienza University of Rome, Italy\\
\email{zanchetta.1848878@studenti.uniroma1.it, amerini@diag.uniroma1.it, maiano@diag.uniroma1.it}\\
 \and
ESA philab\\
\email{lorenzo.papa@esa.int}}

\maketitle

\begin{abstract}

Text-to-video generation is an emerging field in generative AI, enabling the creation of realistic, semantically accurate videos from text prompts. While current models achieve impressive visual quality and alignment with input text, they typically rely on static knowledge—making it difficult to incorporate new data without retraining from scratch. To address this limitation, we propose \textbf{VidCLearn}, a continual learning framework for diffusion-based text-to-video generation. \textbf{VidCLearn} features a student-teacher architecture where the student model is incrementally updated with new text-video pairs, and the teacher model helps preserve previously learned knowledge through generative replay. Additionally, we introduce a novel temporal consistency loss to enhance motion smoothness and a video retrieval module to provide structural guidance at inference. Our architecture is also designed to be more computationally efficient than existing models retaining a satisfactory generation performance. Experimental results show \textbf{VidCLearn} superiority over baseline methods in terms of visual quality, semantic alignment, and temporal coherence.

\end{abstract}

\section{Introduction}
\label{sec:intro}
Text-to-Video (T2V) generation is a cutting-edge computer vision task that involves generating dynamic video sequences starting from a textual description; it aims at filling the existing gap between natural language understanding and video synthesis. Key aspects of this task are the following: \textit{i)} semantic alignment of generated videos with respect to the input textual description; \textit{ii)} temporal consistency between generated frames, and \textit{iii)} high-quality generated frames, possibly avoiding visual artifacts.
However, text-to-video generative models have to face several challenges, including the need for high-quality large-scale datasets, huge computational resources and a large amount of time to ensure a proper training. Another crucial limitation of existing text-to-video generative models is the following: their knowledge is \textit{static}, in the sense that it's not easy to maintain the knowledge of such models up to date, because it would require a new training from scratch process each time, resulting in a huge waste of time and computation. 

In order to overcome these limitations, we propose \textbf{VidCLearn}: a generative replay continual learning approach for training a text-to-video generative model in such a way that its knowledge increases over time. Our work builds upon Tune-A-Video \cite{b1}, a text-to-video generative model that is able to generate a novel video starting from a single text-video pair as input and an edited prompt that guides the generation. Tune-A-Video preserves the motion information at inference time thanks to a DDIM inversion operation performed on the input video, which produces latent noisy frames providing structural guidance during the denoising procedure. Our proposed method consists of a knowledge distillation-like paradigm, where a \textit{student} model is fine-tuned on the currently available text-video pair, while a \textit{teacher} model is in charge of preserving as much as possible the student's previously acquired knowledge over time. Additionally, in order to provide structural guidance to the model at inference time, we propose to integrate a \textit{retrieval mechanism} based on a cosine similarity score computed between the inference prompt and all the training prompts the model has seen so far. Our experiments demonstrate the superiority of the proposed approach with respect to the vanilla Tune-A-Video model and several baseline strategies, which lack of generalization capabilities. We achieve encouraging results that prove the effectiveness of our approach, without significantly increasing the computational requirements for proper training. In fact, our approach leverages only one NVIDIA A100 GPU, resulting in minimal energy consumption and resource-intensive computations, thanks to the almost lightweight core Tune-A-Video 
architecture. In summary, we propose a continual learning framework for text-to-video generation, introducing a temporal consistency loss for motion coherence, a cosine similarity-based retrieval mechanism for structural guidance, and demonstrate its effectiveness through extensive experiments.


\section{Related Works}
The Text-to-Video generation field has evolved rapidly in the past few years, addressing complex challenges like temporal and spatial coherence of the generated video frames. Models like \textit{Sync-DRAW} \cite{b2} leverage attention mechanisms for enhancing temporal consistency, while other models like \textit{ModelScopeT2V} \cite{b5} propose to adapt a pre-trained Stable Diffusion model \cite{b6} to the temporal domain. The \textit{Video Diffusion Model} proposed by Ho et al. \cite{b7} employs diffusion models for generating videos in the pixel space, while other models like \textit{PixelDance} \cite{b8} and \textit{MagicVideo} \cite{b9} propose to generate videos leveraging the latent space, which is less costly. Hybrid models, such as \textit{Show-1} \cite{b11}, combine pixel-based and latent-based video diffusion models, in order to exploit the advantages of both approaches while reducing their drawbacks. There exist also text-to-video generative models based on autoregressive approaches, such as \textit{NUWA-Infinity} \cite{b13}, \textit{CogVideo} \cite{b14}, and \textit{Phenaki} \cite{b15}. Transformer-based approaches, such as \textit{NUWA} \cite{b16} and \textit{GenTron} \cite{b17}, are able to generate high-quality videos, too. The current state-of-the-art in text-to-video synthesis is represented by \textit{Sora} \cite{b18}, a diffusion transformer-based model that has demonstrated a remarkable ability to generate even one minute-long videos while maintaining high visual quality, unlike earlier works, that are able to generate only short clips. Efficient models such as \textit{VideoLCM} \cite{b19}, which is able to achieve high-fidelity video synthesis with only $4 \thicksim 6$ sampling steps, or \textit{Text2Video-Zero} \cite{b20}, which is able to leverage the power of existing text-to-image synthesis methods in order to make them suitable for the video domain in a “zero-shot” fashion, allow to perform video generation from text with significantly reduced computational cost. There exist also efficient models that are able to significantly limit the amount of needed training data, such as \textit{Make-A-Video} \cite{b21} or the previously mentioned \textit{Tune-A-Video}, that have revolutionized the text-to-video generation field.

Continual learning strategies in generative models, particularly diffusion models, address the challenge of catastrophic forgetting. Most of them are based on a generative replay approach; among them, \textit{GPPDM} \cite{b22} leverages the concept of class prototype in order to capture the core characteristics of images belonging to a given class. A key contribution is that of \textit{DSG} \cite{b23} and \textit{Generative Distillation} \cite{b24} models, which employ knowledge distillation for transferring the knowledge in an effective way; the strategies proposed in these two works have inspired our proposed method. Other approaches, such as \textit{STAMINA} \cite{b25} and \textit{C-LoRA} \cite{b26}, involve low-rank adaptation mechanisms for enhancing knowledge retention. There exist also class incremental learning approaches, such as \textit{DiffClass} \cite{b27}, trying to overcome limitations like privacy issues or domain gaps between synthetic data and real data by means of a multi-distribution-matching technique for fine-tuning diffusion models.


Continual learning, despite its potential to enable dynamic knowledge expansion, has been largely unexplored in text-to-video generation. In this work, we take a step toward addressing this gap by proposing a generative replay-based continual learning strategy for text-to-video synthesis.

\section{Methodology}

\begin{figure*}[htbp]
\centerline{\includegraphics[width=9cm]{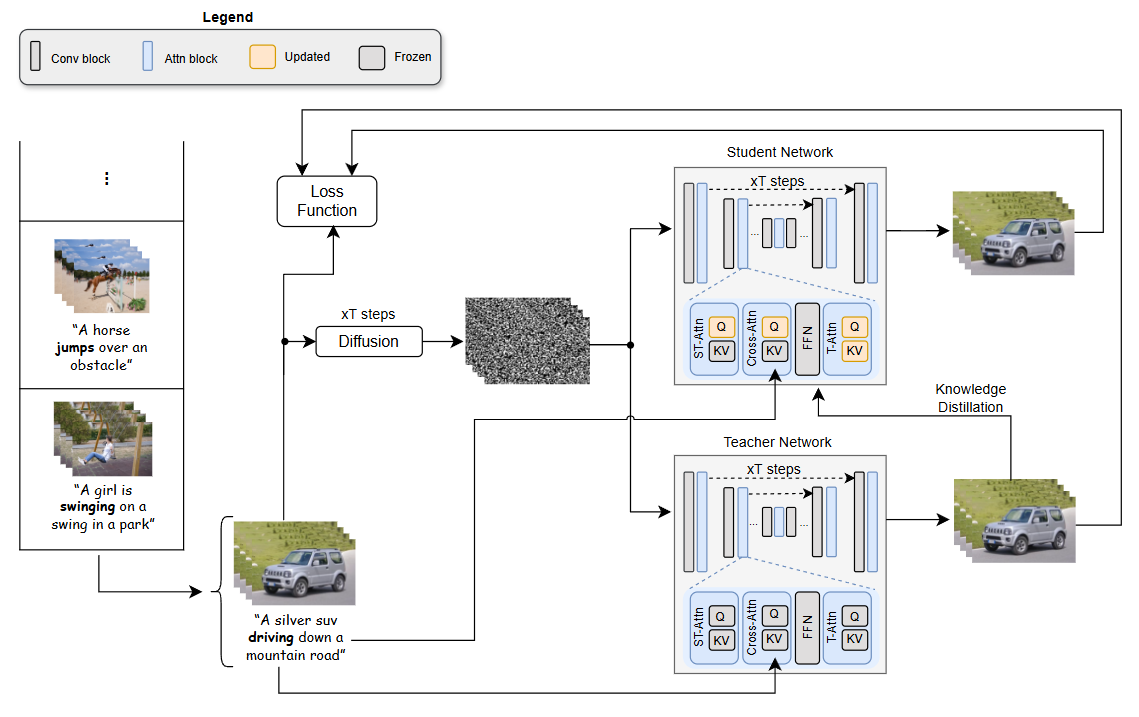}}
\caption{VidCLearn pipeline.} 
\label{fig:vidclearn}
\end{figure*}

Consider a setting in which text-video pairs become available over time, one at a time. Our goal is to train a text-to-video generative model in such a way that its knowledge can be updated over time, without the need to re-train the model from scratch each time a new text-video pair becomes available. 

\begin{figure}[!t]
\centerline{\includegraphics[width=6cm]{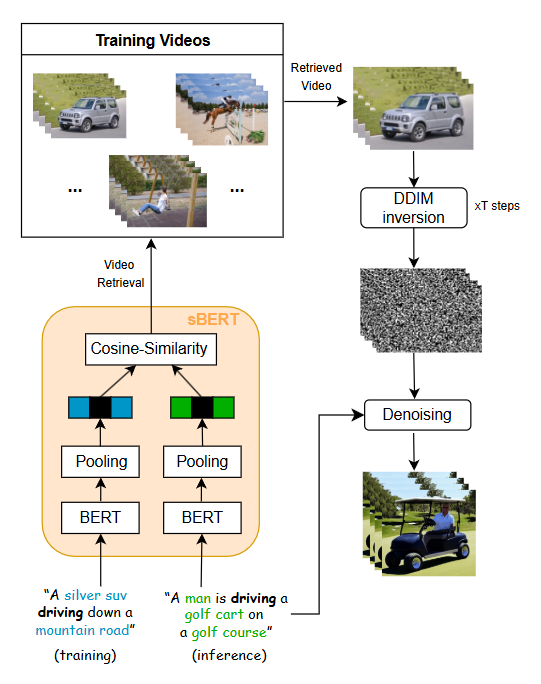}}
\caption{VidCLearn inference procedure.}
\label{fig:inference}
\end{figure}

We propose \textbf{VidCLearn}: a novel approach employing a continual learning technique for text-to-video generation. Specifically, our strategy consists of applying a generative replay continual learning approach to the core Tune-A-Video model, so that its knowledge is both preserved and increased over time. We achieve this by means of a knowledge distillation-like paradigm: the core Tune-A-Video model acts as the student, while a duplicate of the student with frozen weights acts as the teacher. The pipeline of our approach is shown in Figure \ref{fig:vidclearn}. When the very first text-video pair becomes available, there is no prior knowledge to retain; therefore, the student model is tuned by means of the one-shot video tuning strategy of the core Tune-A-Video model. In particular, a latent representation of the input video is created by means of the underlying Stable Diffusion's variational autoencoder; then, a standard forward diffusion process is performed on the latent representation of each video frame, thus producing noisy latents. During the reverse diffusion process, the student model is asked to predict the amount of noise to be removed from each noisy latent representation, in order to get back the original ones; a target amount of noise is defined based on the noise scheduler of the model, and a reconstruction loss is computed
at each training step between the prediction of the model and the predefined target amount of noise. As soon as a new text-video pair becomes available, before tuning is actually performed, the student model is duplicated, and the weights of the duplicate are frozen; this ensures that, during the training process, the duplicate retains the knowledge the student has acquired so far. The student model is now free to learn; the original one-shot video tuning strategy is applied once again, but now the duplicate contributes to the learning process of the student by means of a \textit{distillation loss}, essentially acting as a teacher. First of all, a Mean Squared Error (MSE) is computed between the output of the student network (i.e., the amount of noise to be removed from the current denoising step) $\hat{y}_i$ and the target noise $y_i$ to be removed for the $i$-th element in the noisy video sequence:
\begin{equation}
    L_s = \frac{1}{N} \sum_{i=1}^{N} \left(y_i - \hat{y}_i\right)^2.
\end{equation}
Here, $N$ denotes the total number of frames, while $L_s$ denotes the student's reconstruction loss. The teacher model is then asked to distill its knowledge by means of a KL-Divergence loss:
\begin{equation}
    L_{KL} = T^2 \cdot \infdiv*{Softmax\left(\frac{\textbf{t}}{T}\right)}{ \log \left( Softmax\left(\frac{\textbf{s}}{T}\right) \right)}.
\end{equation}
Here, $T$ is a temperature parameter used to control the softness of the output distributions \textbf{t} and \textbf{s} produced by the teacher and the student networks, respectively. The distillation loss can be thus obtained by means of a linear combination between the two terms $L_{KL}$ and $L_s$ \cite{b29}:
\begin{equation}
\label{eq:distillation}
    L_{distill} = \alpha \cdot L_{KL} + (1 - \alpha) \cdot L_s,
\end{equation}
where $\alpha$ is a crucial parameter that balances knowledge retention with the learning of new knowledge. Notice that, when the model is being tuned on the very first text-video pair, we have $L_{distill} = L_s$, without any $\alpha$ coefficient. Additionally, in order to enhance the temporal consistency between the generated frames, we propose to integrate a \textit{temporal consistency loss} considering the difference on the temporal dimension between consecutive frames for both the prediction of the student model and the target noise that should be predicted. In particular, let us denote with $\boldsymbol{P} \in \mathbb{R}^{B \times C \times T \times H \times W}$ the model's prediction, and with $\boldsymbol{N} \in \mathbb{R}^{B \times C \times T \times H \times W}$ the target noise the model's prediction has to resemble. A squared difference between the temporal changes of both the predictions of the model and the target noise, averaged over all the frames $M$, is then computed as follows:
\begin{equation}
\label{eq:tloss}
    L_t = \frac{1}{M} \sum_{i=1}^{M} \left(\left(\boldsymbol{P}_i^{(t+1)} - \boldsymbol{P}_i^{(t)}\right) - \left(\boldsymbol{N}_i^{(t+1)} - \boldsymbol{N}_i^{(t)}\right)\right)^2
\end{equation}
Therefore, considering both the distillation loss and the temporal consistency loss we have previously defined, our final proposed loss is defined as follows:
\begin{equation}
    L_{tot} = \gamma \cdot L_{distill} + \lambda \cdot L_t,
\end{equation}
where $\gamma$ and $\lambda$ are additional coefficients used for scaling the contribution of the two loss components with respect to the total loss function. Additionally, we propose to integrate a \textit{retrieval mechanism} in order to choose the training video that should be used for providing structural guidance to the model at inference time. Our proposed approach, depicted in Figure \ref{fig:inference},
is based on Sentence-BERT (sBERT) \cite{b30}: a modification of the BERT model \cite{b31} using siamese and triplet networks that is able to derive semantically meaningful sentence embeddings from input sentences. We feed sBERT with the inference prompt and each prompt used for training the model, one by one, and we let sBERT compute a similarity score between them. The highest similarity score is then used to retrieve the training prompt resembling the most the inference prompt, and the corresponding training video will in turn be used for providing structural guidance to the model at inference time. Finally, a \textit{DDIM inversion} operation is performed on the retrieved video, resulting in an inverted latent noise, that will in turn be used as the starting point for a \textit{DDIM sampling} operation. The latent representation resulting from this latter DDIM sampling operation is then decoded, thus providing the generated video.

\section{Experiments and Results}
In this section, we evaluate the effectiveness of our proposed continual learning approach for text-to-video generative models, with the aim of maximizing temporal consistency and generation quality while minimizing knowledge degradation over time.

\textbf{Dataset.} We evaluate our method on the DAVIS dataset \cite{b32}, as done by the authors of Tune-A-Video. Since the plain DAVIS dataset does not provide paired text-video data, we employ a pre-trained BLIP-2 model \cite{b33} for generating the textual descriptions of each video frame. The caption representing each single video is determined by the highest CLIP Score \cite{b34} value resulting from each group of frames, along with their corresponding descriptions. This setup guarantees the availability of 90 text-video pairs for training and 30 text-video pairs for evaluation, to be provided to the model over time.

\textbf{Evaluation Metrics.} In order to evaluate different aspects of our model's performance, we rely on some of the most commonly used evaluation metrics. We evaluate the visual quality of the model's output by means of the the \textit{Fréchet Video Distance} (FVD) \cite{b35}, the \textit{Fréchet Inception Distance} (FID) \cite{b36}, and the \textit{Inception Score} (IS) \cite{b37} metrics. We also evaluate the semantic alignment between the generated content and the corresponding prompt by means of the \textit{CLIP Score} metric. Finally, we evaluate the knowledge retention's degree of our model by means of the \textit{Forward Transfer} (FWT) and the \textit{Backward Transfer} (BWT) metrics \cite{b38}.

\textbf{Implementation Details.} At training time, we sample 20 uniform frames at resolution of $512\times512$ from each input video, and we fine-tune the student model for 500 steps with a learning rate $\eta$ of $3\times10^{-5}$, a temperature $T$ of 4, a distillation weight $\gamma$ of 1 and a temporal weight $\lambda$ of 10. Moreover, the balancing coefficient $\alpha$ of the distillation loss (equation \ref{eq:distillation}) is set to 0.8. At inference time, we perform DDIM inversion with 100 inverted latent steps, and then we generate the output video with 150 inference steps. Each tuning on the current text-video pair takes slightly more than one hour on an NVIDIA A100 GPU, and requires around 37 MiB of GPU RAM to be performed; at inference time, each generation takes about one minute to be performed and requires around 7 MiB of GPU RAM.

\subsection{Quantitative Evaluation}
\begin{table}[!t]
\caption{Naive Approach: visual quality and semantic alignment.}
\centering
\scriptsize 
\begin{tabular}{|c|c|c|c|c|}
\hline
\textbf{Videos} & \textbf{FVD} $\downarrow$ & \textbf{FID} $\downarrow$ & \textbf{IS} $\uparrow$ & \textbf{CLIP} $\uparrow$ \\
\hline
5 & 865,824 & 9,344 & 10,272 & 32,058 \\
10 & 935,043 & 19,915 & 9,298 & 31,002 \\
25 & 1224,906 & 20,105 & 6,043 & 26,870 \\
50 & 954,725 & 11,570 & 6,891 & 27,677 \\
75 & 952,074 & 11,538 & 7,318 & 28,187 \\
90 & 617,230 & 7,112 & 6,328 & 26,743 \\
\hline
\end{tabular}
\label{tab:baseline}
\end{table}

Since there is no prior work employing continual learning techniques for solving the text-to-video generation task, the reference baseline against which we compare the performance of our method consists on iteratively fine-tuning the Tune-A-Video model on all the training text-video pairs, provided sequentially over time. We measure the behavior of the baseline model over time, as more text-video pairs become available; the results, summarized in Table \ref{tab:baseline}, show an overall decrease of the model's performance, as the number of text-video pairs increases over time. This is a clear sign of the catastrophic forgetting issue arising. 

One of the simplest strategies to avoid the catastrophic forgetting issue is the \textit{Elastic Weight Consolidation (EWC)} technique \cite{b39}: this technique selectively protects important parameters of the model as its knowledge is expanded over time. By applying the EWC technique, as summarized in Table \ref{tab:ewc}, we notice a slight increase in performance with respect to the naive baseline: we obtain a reduction of 15,42\% in FID, an increase of 20,93\% in IS and an increase of 5,46\% in CLIP Score. However, these slight variations show that the behavior of the model remains almost the same as before: the catastrophic forgetting issue still persists, as the performance of the model decreases over time.
\begin{table}[!t]
\caption{EWC Approach: visual quality and semantic alignment.}
\centering
\scriptsize
\begin{tabular}{|c|c|c|c|c|}
\hline
\textbf{Videos} & \textbf{FVD} $\downarrow$ & \textbf{FID} $\downarrow$ & \textbf{IS} $\uparrow$ & \textbf{CLIP} $\uparrow$ \\
\hline
5 & 1096,049 & 10,155 & 10,820 & 31,839 \\
10 & 1210,202 & 25,408 & 9,589 & 30,278 \\
25 & 1593,644 & 18,167 & 6,751 & 28,508 \\
50 & 1264,515 & 12,656 & 6,942 & 28,200 \\
75 & 844,169 & 7,481 & 7,091 & 25,641 \\
90 & 901,032 & 6,162 & 8,003 & 28,288 \\
\hline
\end{tabular}
\label{tab:ewc}
\end{table}

Our proposed approach significantly reduces the catastrophic forgetting issue arising in the baseline, while improving both visual quality and temporal consistency of the generated video frames. Compared to the baseline, as shown in Table \ref{tab:vidclearn}, experiments performed with our proposed method demonstrate an overall improvement of the model's performance. Specifically, since we obtain a 18,97\% increase in FVD, a 125,42\% reduction in FID, and a 102,78\% increase in IS, the visual quality of the generated videos improves. Additionally, since we obtain a 23,03\% increase in CLIP Score, also the semantic alignment between the generated content and the inference prompt improves. We can draw similar conclusions by comparing our approach with the EWC strategy: in this latter case, we obtain a 22,7\% reduction in FVD, a 95,31\% reduction in FID and a 37,63\% increase in IS, demonstrating better performance in terms of visual quality of generated videos. In addition, we also obtain a 14,02\% increase in the CLIP score, showing an improvement in performance also in terms of semantic alignment.
\begin{table}[!t] 
\caption{VidCLearn Approach: visual quality and semantic alignment.}
\scriptsize
\centering
\begin{tabular}{|c|c|c|c|c|} 
\hline 
\textbf{Videos} & \textbf{FVD} $\downarrow$ & \textbf{FID} $\downarrow$ & \textbf{IS} $\uparrow$ & \textbf{CLIP} $\uparrow$ \\ 
\hline 
5 & 740,210 & 8,505 & 12,002 & 32,800 \\ 
10 & 820,430 & 15,704 & 10,885 & 31,754 \\  
25 & 980,112 & 14,002 & 9,123 & 29,932 \\  
50 & 852,764 & 10,200 & 9,005 & 30,125 \\  
75 & 765,821 & 6,512 & 10,304 & 29,342 \\ 
90 & 734,320 & 3,155 & 12,832 & 32,903 \\ 
\hline 
\end{tabular}
\label{tab:vidclearn}
\end{table}

Our proposed approach also shows a significant performance improvement in terms of knowledge retention. In fact, when compared with the Naive approach, our proposed VidCLearn approach leads to a 215,17\% increase in FWT (1,002 Naive vs 3,158 VidCLearn) and a 158,73\% increase in BWT (0,538 Naive vs 1,392 VidCLearn); moreover, when compared with the EWC strategy, our approach leads to a 30,62\% improvement in FWT (2,191 EWC vs 3,158 VidCLearn) and a 27,23\% improvement in BWT (1,013 EWC vs 1,392 VidCLearn). All the experiments were conducted after the knowledge of the model was expanded with 90 training videos.

\subsection{Qualitative Evaluation}
\begin{figure}[!t]
\centerline{\includegraphics[width=8cm]{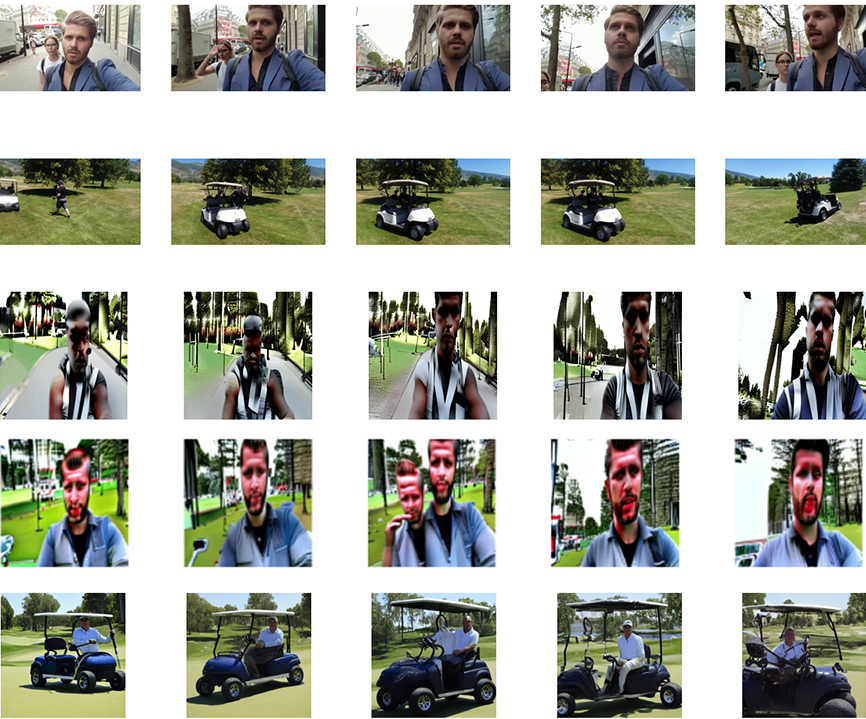}}
\caption{Qualitative evaluation: first row shows the frames of the last training video; second row shows the ground truth with the textual prompt “\textit{a man is driving a golf cart on a golf course}”; third row shows the video frames generated by the Naive approach, the fourth with the EWC approach and the last row with the proposed \textbf{VidCLearn} approach.}
\label{fig:qualitative}
\end{figure}

Table \ref{tab:baseline} shows, from a quantitative point of view, that iteratively fine-tuning the Tune-A-Video model on all the training text-video pairs that become available sequentially over time leads to catastrophic forgetting. Let's see what happens visually when we ask the model to generate a video in response to the inference prompt “\textit{a man is driving a golf cart on a golf course}”; Figure \ref{fig:qualitative} shows its behavior. Specifically, the first row shows the last training video on which the model was tuned; the second row, instead, shows the ground truth video, while the last three rows show the actual generated video with the different methodologies. In the third row we can see that with the Naive approach the model responds to the inference prompt by generating a video that highly resembles the last video seen during training.
A very similar behavior is observed when applying the \textit{Elastic Weight Consolidation (EWC)} technique: quantitatively, the performance of the model decreases over time, despite a slight overall increase with respect to the naive baseline. Let's see again what happens visually when we ask the model to generate a video in response to the same inference prompt as before; Figure \ref{fig:qualitative} (see row number four) shows its behavior. Even the application of the EWC technique results in the model generating a video that highly resembles the last video seen during training, in response to the given inference prompt; however, in this case, the visual quality of the generated frames appears to be better than before.
Table \ref{tab:vidclearn}, on the other hand, shows that the application of our proposed approach leads to a significant increase in the performance of the model: the generated video frames have better visual quality and temporal consistency, and the degree of knowledge retention is higher. Let's see again what happens visually when we ask the model to generate a video in response to the same inference prompt as before; Figure \ref{fig:qualitative} (last row) shows its behavior. With our approach, the model responds to the inference prompt by generating a video that highly resembles the corresponding ground truth video, as an additional proof of the superiority of our proposed method.

\subsection{Ablation Study}
To evaluate the contribution of each component in our proposed approach, we have conducted an ablation study by systematically removing or altering individual components, and analyzing their impact on model performance. We evaluate the performance of our model after learning the entire training partition of the DAVIS dataset using metrics such as \textit{FVD}, \textit{FID}, \textit{IS}, and \textit{CLIP score}. 
Therefore, we investigated the impact of the temporal consistency loss (equation \ref{eq:tloss}) on model performance by increasing the temporal importance factor $\lambda$. Table \ref{tab:abltemporal} summarizes the results of our experiments. Clearly, the presence of this component benefits the overall performance of the model; however, the temporal importance factor $\lambda$ should not be set too high, otherwise performance starts to degrade. For instance, as reported in Table \ref{tab:abltemporal}, we can see that as the temporal importance factor $\lambda$ increases, the performance of the model in terms of the semantic alignment between its input and its output decreases: in fact, the CLIP score decreases from 32,903 with $\lambda = 10$, to 32,783 with $\lambda = 20$, to 32,651 with $\lambda = 30$.
\begin{table}[!t]
\caption{Ablation study on the temporal loss.}
\scriptsize
\centering
\begin{tabular}{|c|c|c|c|c|c|c|}
\hline
\textbf{$\alpha$} & \textbf{$\gamma$} & \textbf{$\lambda$} & \textbf{FVD} $\downarrow$ & \textbf{FID} $\downarrow$ & \textbf{IS} $\uparrow$ & \textbf{CLIP} $\uparrow$ \\
\hline
0,7 & 1 & 0 & 791,997 & 4,007 & 12,613 & 32,699 \\
0,7 & 1 & 1 & 774,002 & 4,249 & 12,437 & 32,771 \\
0,8 & 1 & 10 & 734,320 & 3,155 & 12,832 & 32,903 \\
0,8 & 1 & 20 & 699,161 & 3,051 & 12,165 & 32,783 \\
0,8 & 1 & 30 & 660,582 & 3,090 & 13,217 & 32,651 \\
\hline
\end{tabular}
\label{tab:abltemporal}
\end{table}

\begin{table}[!t]
\caption{Ablation study on the retrieval mechanism.}
\scriptsize
\centering
\begin{tabular}{|c|c|c|c|c|c|c|c|c|}
\hline
\textbf{Guidance} & \textbf{$\alpha$} & \textbf{$\gamma$} & \textbf{$\lambda$} & \textbf{FVD} $\downarrow$ & \textbf{FID} $\downarrow$ & \textbf{IS} $\uparrow$ & \textbf{CLIP} $\uparrow$ \\
\hline
Last & 0,7 & 1 & 0 & 782,928 & 3,541 & 10,874 & 31,490 \\
Retrieval & 0,7 & 1 & 0 & 791,997 & 4,007 & 12,613 & 32,699 \\
\hline
Last & 0,7 & 1 & 1 & 758,124 & 3,322 & 10,832 & 31,631 \\
Retrieval & 0,7 & 1 & 1 & 774,002 & 4,249 & 12,437 & 32,771 \\
\hline
Last & 0,8 & 1 & 10 & 728,164 & 2,899 & 11,324 & 31,898 \\
Retrieval & 0,8 & 1 & 10 & 734,320 & 3,155 & 12,832 & 32,903 \\
\hline
Last & 0,8 & 1 & 20 & 709,237 & 3,111 & 11,200 & 32,080 \\
Retrieval & 0,8 & 1 & 20 & 699,161 & 3,051 & 12,165 & 32,783 \\
\hline
Last & 0,8 & 1 & 30 & 684,854 & 3,169 & 10,871 & 32,159 \\
Retrieval & 0,8 & 1 & 30 & 660,582 & 3,090 & 13,217 & 32,651 \\
\hline
\end{tabular}
\label{tab:ablretrieval}
\end{table}

We also investigated the impact of our proposed retrieval mechanism for structural guidance on model performance. Table \ref{tab:ablretrieval} summarizes the results of our experiments; here, “\textit{last}” means that at inference time, structural guidance was provided by the last training video, while “\textit{retrieval}” means that structural guidance was provided by the training video retrieved by our proposed retrieval mechanism. Our model benefits from the dynamic structural guidance provided by such a retrieval mechanism, leading to an overall increase in model performance.

\section{Conclusions}
This work presents a continual learning approach for text-to-video generation, addressing issues like catastrophic forgetting and motion consistency. Existing models are almost static and cannot adapt to new data without full retraining. We tackle this by applying continual learning techniques to enable dynamic knowledge expansion over time. To reduce computational costs, we adopt a lightweight architecture inspired by Tune-A-Video. Our method achieves promising results, preserving prior knowledge while maintaining spatial and temporal coherence. However, challenges remain, including difficulty generating unseen motion patterns and improving visual quality. Future work could explore perceptual losses, memory replay strategies, and model compression techniques like pruning or quantization to address these limitations more efficiently.

\section{Acknowledgments}
This study has been partially supported by SERICS (PE00000014) under the MUR National Recovery and Resilience Plan funded by the European Union - NextGenerationEU.

%
%
\bibliographystyle{splncs04}

\end{document}